\documentclass{article}

\usepackage{graphicx}
\usepackage{amssymb}

\begin{document}


\title{Topological Grammars for Data Approximation}

\author{A.N. Gorban$^{1,3}$\thanks{ag153@le.ac.uk}, N.R.
Sumner$^{1}$\thanks{nrs7@le.ac.uk}, and A.Y.
Zinovyev$^{2,3}$\thanks{andrei.zinovyev@curie.fr} \\
 $ ^{1}$University of Leicester, Leicester, LE1 7RH, UK  \\
 $ ^{2}${Institut Curie, rue d'Ulm, Paris, 75248, France} \\
 $ ^{3}$Institute of Computational Modeling SB RAS, Krasnoyarsk, Russia}

\date{}

\maketitle

\begin{abstract}
A method of {\it topological grammars} is proposed for
multidimensional data approximation. For data with complex topology
we define a {\it principal cubic  complex} of low dimension and
given complexity that gives the best approximation for the dataset.
This complex is a generalization of linear and non-linear principal
manifolds and includes them as particular cases. The problem of
optimal principal complex construction is transformed into a series
of minimization problems for quadratic functionals. These quadratic
functionals have a physically transparent interpretation in terms of
elastic energy. For the energy computation, the whole complex is
represented as a system of nodes and springs. Topologically, the
principal complex is a product of one-dimensional continuums
(represented by graphs), and the grammars describe how these
continuums transform during the process of optimal complex
construction. This factorization of the whole process onto
one-dimensional transformations using minimization of quadratic
energy functionals allow us to construct efficient algorithms.
\end{abstract}


\section{Introduction}

In this paper, we discuss a classical problem: how to approximate a
finite set $D$ in $R^m$ for relatively large $m$ by a finite subset
of a regular low-dimensional object in $R^m$. In application, this
finite set is a dataset, and this problem arises in many areas: from
data visualization to fluid dynamics.

The first hypothesis we have to check is: whether the dataset $D$ is
situated near a low--dimensional affine manifold (plane) in $R^m$.
If we look for a point, straight line, plane, ... that minimizes the
average squared distance to the datapoints, we immediately come to
the Principal Component Analysis (PCA). PCA is one of the most
seminal inventions in data analysis. Now it is textbook material.
Nonlinear generalization of PCA is a great challenge, and many
attempts have been made to answer it. Two of them are especially
important for our consideration: Kohonen's Self-Organizing Maps
(SOM) and principal manifolds.

With the {\it SOM} algorithm \cite{Kohonen82} we take a finite
metric space $Y$ with metric $\rho$ and try to map it into $R^m$
with (a) the best preservation of initial structure in the image of
$Y$ and (b) the best approximation of the dataset $D$. The  SOM
algorithm has several setup variables to regulate the compromise
between these goals. We start from some initial approximation of the
map, $\phi_1 : M \rightarrow R^m$. On each ($k$-th) step of the
algorithm we have a datapoint $x \in D$ and a current approximation
$\phi_k : M \rightarrow R^m$. For these $x$ and $\phi_k$ we define
an ``owner" of $x$ in $Y$: $y_x = {\rm argmin}_{y \in Y} \|
x-\phi_k(y) \|$. The next approximation, $\phi_{k+1}$, is
\begin{equation}\label{SOM}
\phi_{k+1}(y) =\phi_{k}(y)+ h_k w(\rho(y,y_x))( x-\phi_k(y)).
\end{equation}
Here $h_k$ is a step size, $0\leq w(\rho(y,y_x))\leq 1$ is a
monotonically decreasing cutting function. There are many ways to
combine steps (\ref{SOM}) in the whole algorithm. The idea of SOM
is very flexible and seminal, has plenty of applications and
generalizations, but, strictly speaking, we don't know what we are
looking for: we have the algorithm, but no independent definition:
SOM is a result of the algorithm work. The attempts to define SOM
as solution of a minimization problem for some energy functional
were not very successful \cite{Erwin92}.

For a known probability distribution, {\it principal manifolds} were
introduced as lines or surfaces passing through ``the middle'' of
the data distribution \cite{HastieStuetzle89}. This intuitive vision
was transformed into the mathematical notion of {\it
self-consistency}: every point $x$ of the principal manifold $M$ is
a conditional expectation of all points $z$ that are projected into
$x$. Neither manifold, nor projection should be linear: just a
differentiable projection $\pi$ of the data space (usually it is
$R^m$ or a domain in $R^m$) onto the manifold $M$ with the
self-consistency requirement for conditional expectations: $
x=\mathbf{E}(z|\pi(z)=x).$ For a finite dataset $D$, only one or
zero datapoints are typically projected into a point of the
principal manifold. In order to avoid overfitting, we have to
introduce smoothers that become an essential part of the principal
manifold construction algorithms.

SOMs give the most popular approximations for principal manifolds:
we can take for $Y$ a fragment of a regular $k$-dimensional grid and
consider the resulting SOM as the approximation to the
$k$-dimensional principal manifold (see, for example,
\cite{Mulier95,Ritter92}). Several original algorithms for
construction of principal curves \cite{Kegl02} and surfaces for
finite datasets were developed during last decade, as well as many
applications of this idea. In 1996, in a discussion about SOM at the
5th Russian National Seminar in Neuroinformatics, a method of
multidimensional data approximation based on elastic energy
minimization was  proposed (see
\cite{GorbanRossiev99,ZinovyevBook00,GorZinComp2005} and the
bibliography there). This method is based on the analogy between the
principal manifold and the elastic membrane (and plate). Following
the metaphor of elasticity, we introduce two quadratic smoothness
penalty terms. This allows one to apply  standard minimization of
quadratic functionals (i.e., solving a system of linear algebraic
equations with a sparse matrix).

\section{Graph grammars and principal graphs}

Let $G$ be a simple undirected graph with set of vertices $Y$ and
set of edges $E$. For $k \geq 2$ a $k$-star in $G$ is a subgraph
with $k+1$ vertices $y_{0,1, \ldots k} \in Y$ and $k$ edges $\{(y_0,
y_i) \ | \ i=1,\ldots k\} \subset E$. Suppose for each $k\geq 2$, a
family $S_k$ of $k$-stars in $G$ has been selected. We call a graph
$G$ with selected families of $k$-stars $S_k$ an {\it elastic graph}
if, for all $E^{(i)} \in E $ and $S^{(j)}_k \in S_k$, the
correspondent elasticity moduli $\lambda_i > 0$ and $\mu_{kj}
> 0$ are defined. Let  $E^{(i)}(0),E^{(i)}(1)$ be vertices of an
edge $E^{(i)}$ and $S^{(j)}_k (0),\ldots S^{(j)}_k (k)$ be vertices
of a $k$-star  $S^{(j)}_k $ (among them, $S^{(j)}_k (0)$ is the
central vertex).
 For any map $\phi:Y \to R^m$ the {\it energy of the
graph} is defined as
\begin{eqnarray}
U^{\phi}{(G)}&:=&\sum_{E^{(i)}} \lambda_i
\left\|\phi(E^{(i)}(0))-\phi(E^{(i)}(1)) \right\| ^2 \\ &&+
\sum_{S^{(j)}_k}\mu_{kj} \left\|\sum _ {i=1}^k \phi(S^{(j)}_k
(i))-k\phi(S^{(j)}_k (0)) \right\|^2. \nonumber
\end{eqnarray}

Very recently, a simple but important fact was noticed
\cite{Gusev04}: every system of elastic finite elements could be
represented by a system of springs, if we allow some springs to have
negative elasticity coefficients.  The energy of a $k$-star $s_k$ in
$R^m$ with $y_0$ in the center and $k$ endpoints $y_{1,\ldots k}$ is
$u_{s_k}= \mu_{s_k}(\sum_{i=1}^k y_i - k y_0)^2$, or, in the spring
representation, $u_{s_k}=k\mu_{s_k} \sum_{i=1}^k (y_i - y_0)^2 -
\mu_{s_k} \sum_{i
> j} (y_i-y_j)^2$. Here we have $k$ positive springs with
coefficients $k\mu_{s_k}$ and $k(k-1)/1$ negative springs with
coefficients $-\mu_{s_k}$.

For a given map $\phi: Y \to R^m$ we divide the dataset $D$ into
subsets $K^y, \, y\in Y$. The set $K^y$ contains the data points for
which the node $\phi(y)$ is the closest one in $\phi(Y)$. The {\it
energy of approximation} is:
\begin{equation}
U^{\phi}_A(G,D):= \sum_{y \in Y} \sum_{ x \in K^y} w(x) \|x-
\phi(y)\|^2,
\end{equation}
where $w(x) \geq 0$ are the point weights.

The simple algorithm for minimization of the energy
$U^{\phi}=U^{\phi}_A(G,D)+U^{\phi}{(G)}$ is the splitting algorithm,
in the spirit of the classical $k$-means clustering: for a given
system of sets $\{K^y \ | \ y \in Y \}$ we minimize $U^{\phi}$ (it
is the minimization of a positive quadratic functional), then for a
given $\phi$ we find new $\{K^y\}$, and so on; stop when no change.
This algorithm gives a local minimum, and the global minimization
problem arises. There are many methods for improving the situation,
but without guarantee of the global minimization.

\begin{figure}[t]
\centering{
\includegraphics[width=37mm, height=35mm]{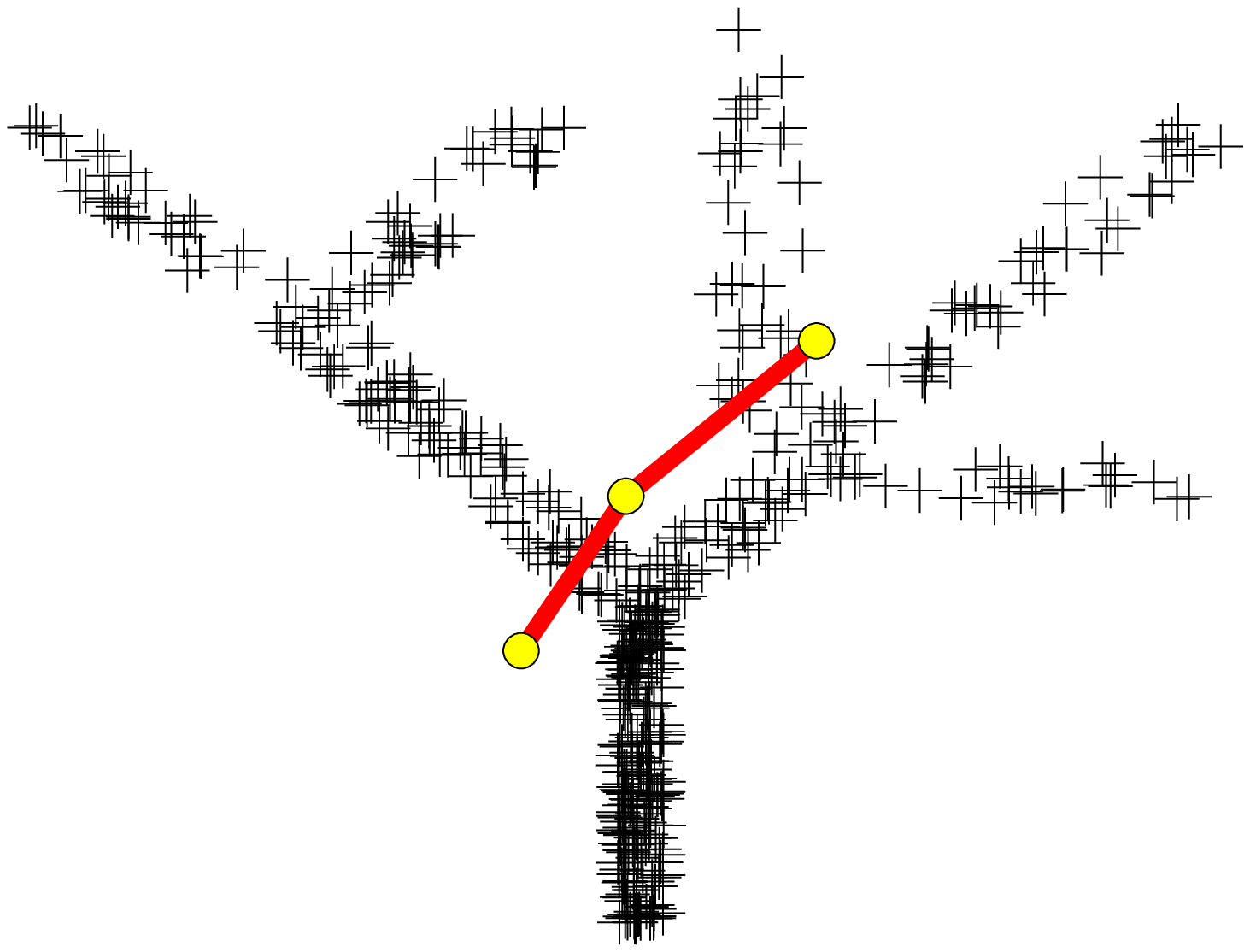}
\includegraphics[width=37mm, height=35mm]{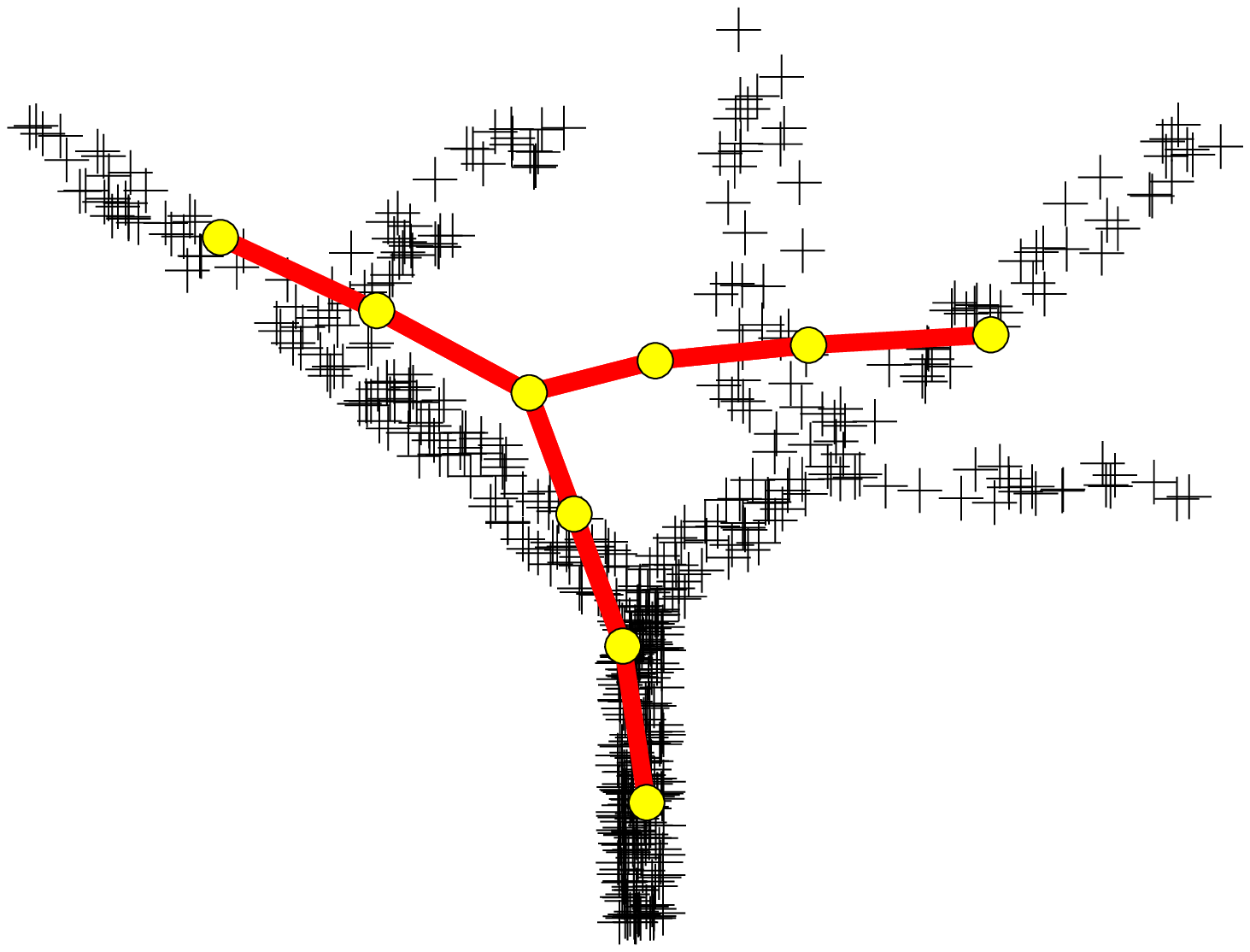}
\includegraphics[width=37mm, height=35mm]{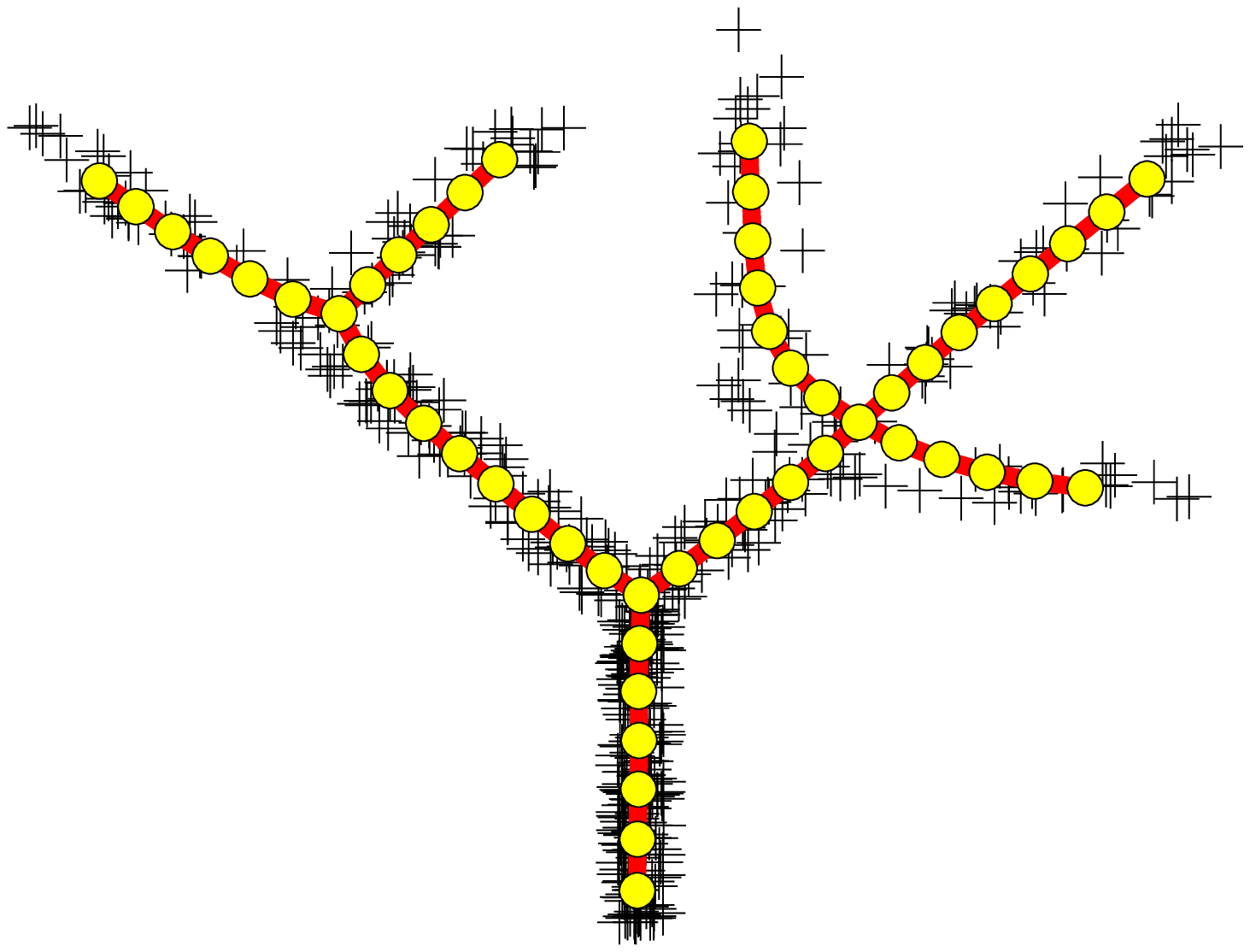}
\\
Iteration 1 \hspace{2.2cm} Iteration 5 \hspace{2.2cm} Iteration 50
\\
\includegraphics[width=37mm, height=35mm]{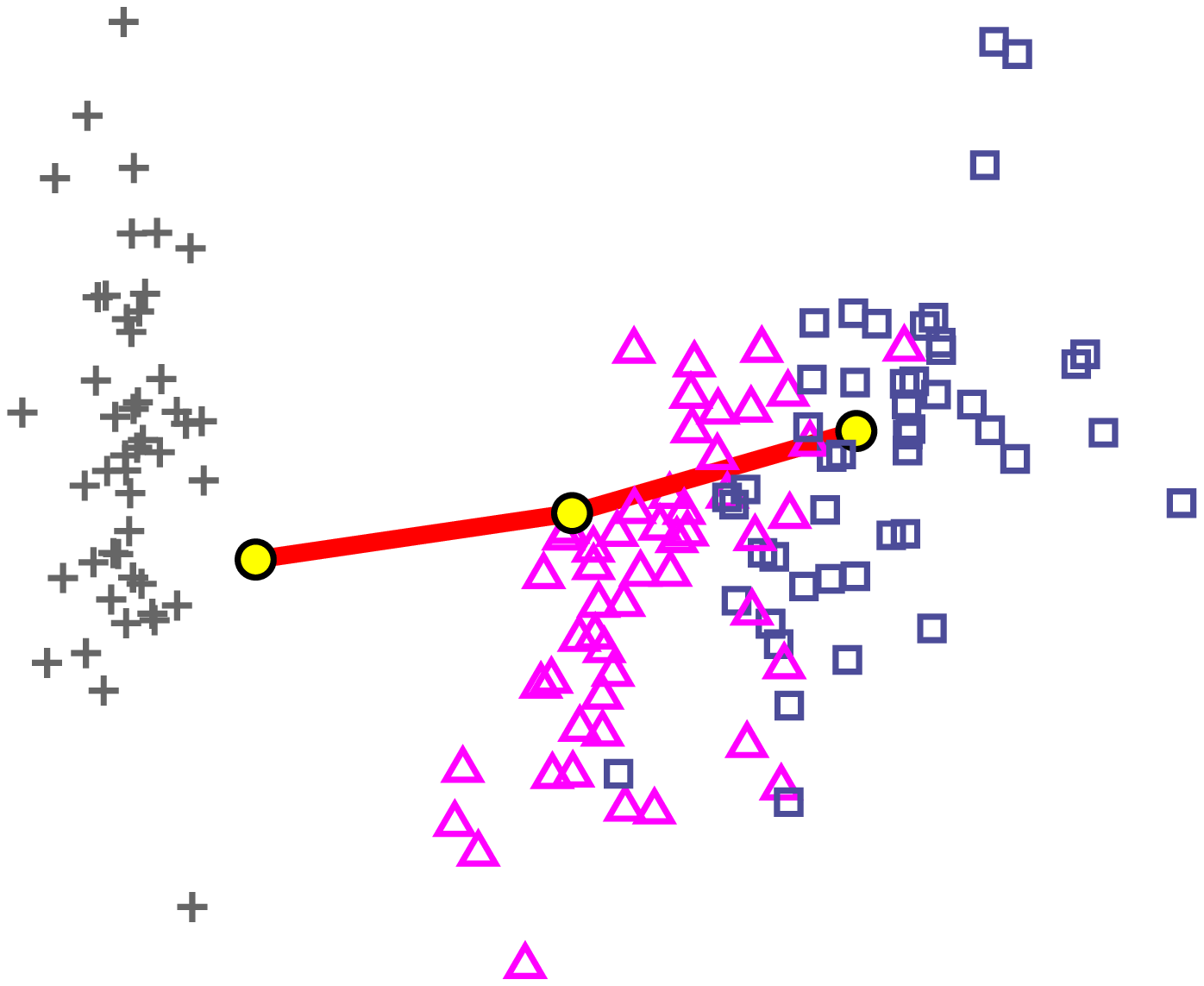}
\includegraphics[width=37mm, height=35mm]{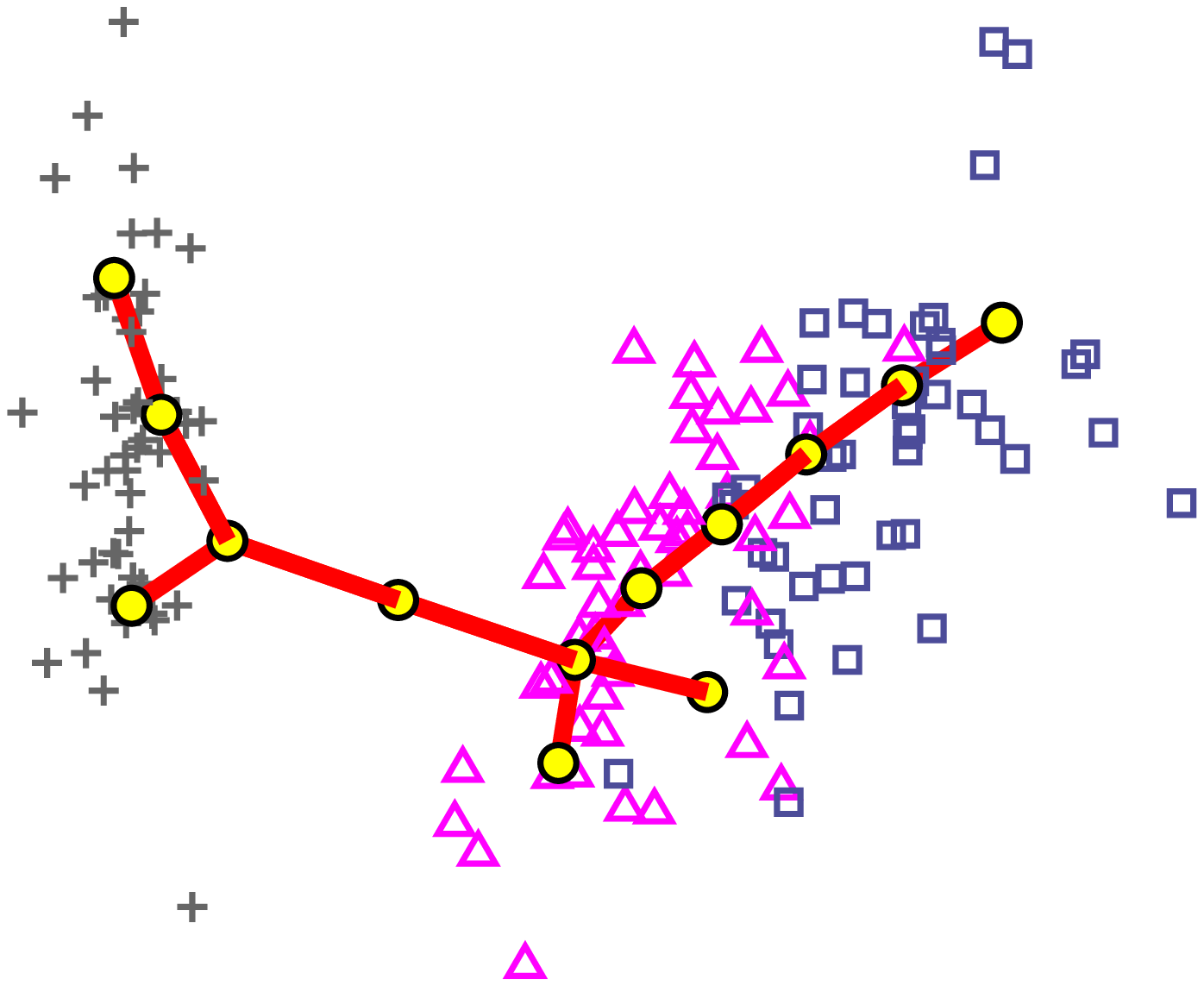}
\includegraphics[width=37mm, height=35mm]{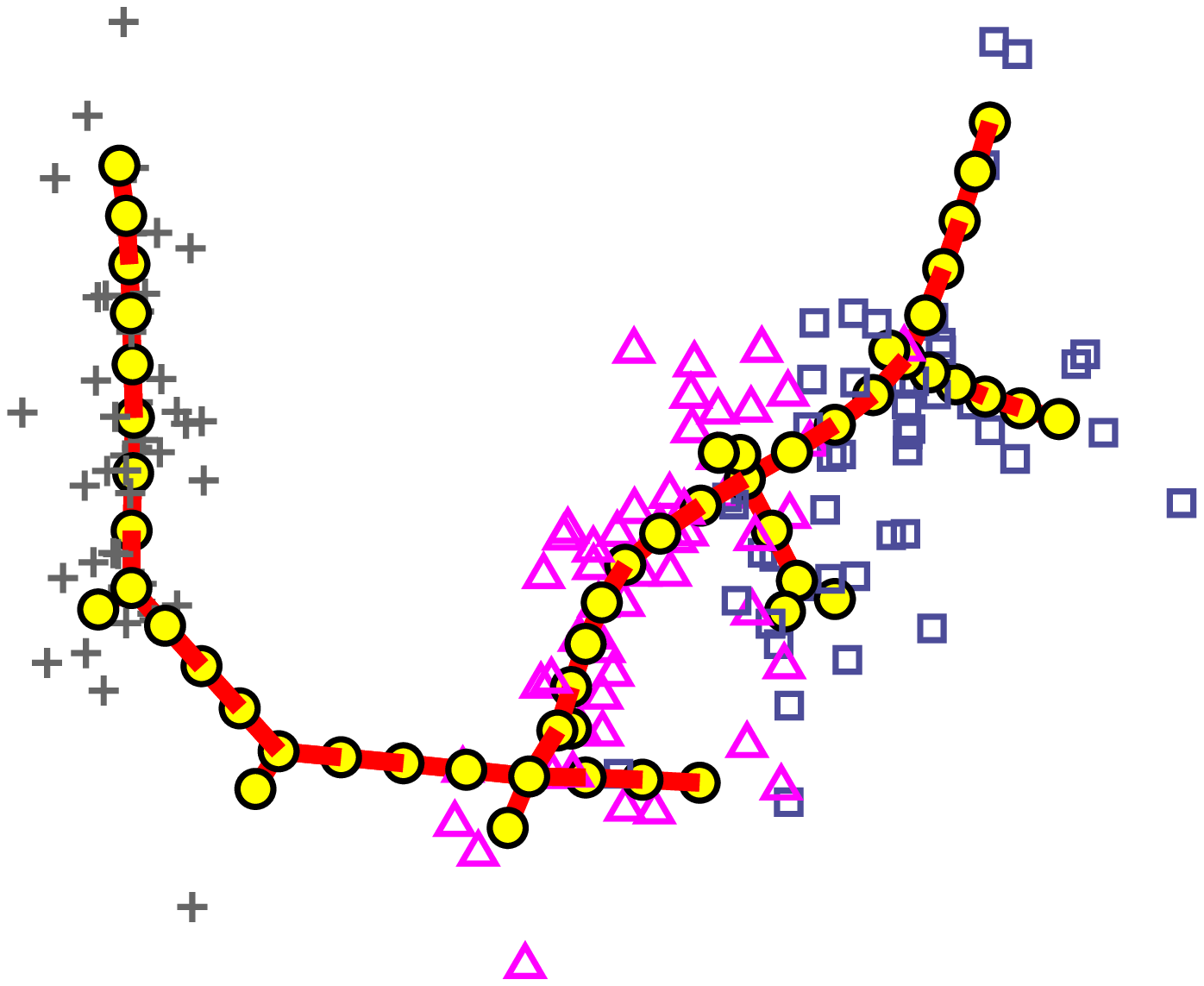}
\\
Iteration 1 \hspace{2.2cm} Iteration 10 \hspace{2.2cm} Iteration 50
} \caption{\label{examples}Applying a simple ``add a node to a node
or bisect an edge" grammar to construct principal elastic trees (one
node is added per iteration). Upper row: an example of
two-dimensional branching distribution of points. Lower row: the
classical benchmark, the  ``iris" four-dimensional dataset (point
shapes distinguish three classes of points), the dataset and
principal tree are presented in projection onto the plane of first
two principal components. }
\end{figure}

The next problem is the elastic graph construction. Here we should
find a compromise between simplicity of graph topology, simplicity
of geometrical form for a given topology, and accuracy of
approximation. Geometrical complexity is measured by the graph
energy $U^{\phi}{(G)}$, and the error of approximation is measured
by the energy of approximation $U^{\phi}_A(G,D)$. Both are
included in the energy $U^{\phi}$. Topological complexity will be
represented by means of elementary transformations: it is the
length of the energetically optimal chain of elementary
transformation from a given set applied to initial simple graph.

Graph grammars \cite{Nagl,Loewe} provide a well-developed formalism
for the description of elementary transformations. An elastic graph
grammar is presented as a set of production (or substitution) rules.
Each rule has a form $A \to B$, where $A$ and $B$ are elastic
graphs. When this rule is applied to an elastic graph, a copy of $A$
is removed from the graph together with all its incident edges and
is replaced with a copy of $B$ with edges that connect $B$ to graph.
For a full description of this language we need the notion of a {\it
labeled graph}. Labels are necessary to provide the proper
connection between $B$ and the graph.

A link in the energetically optimal transformation chain is
constructed by finding a transformation application that gives the
largest energy descent (after an optimization step), then the next
link, and so on, until we achieve the desirable accuracy of
approximation, or the limit number of transformations (some other
termination criteria are also possible). The selection of an
energetically optimal application of transformations by the trial
optimization steps is time-consuming. There exist alternative
approaches. The preselection of applications for a production rule
$A \to B$ can be done through comparison of energy of copies of $A$
with its incident edges and stars in the transformed graph $G$.

As the simple (but already rather powerful) example  we use a system
of two transformations: ``add a node to a node" and ``bisect an
edge." These transformations act on a class of {\it primitive
elastic graphs}:  all non-terminal nodes with $k$ edges are centers
of elastic k-stars, which form all the $k$-stars of the graph. For a
primitive elastic graph, the number of stars is equal to the number
of non-terminal nodes -- the graph topology prescribes the elastic
structure.

The transformation {\it ``add a node"} can be applied to any vertex
$y$ of $G$:  add a new node $z$ and a new edge $(y,z)$. The
transformation {\it ``bisect an edge"} is applicable to any pair of
graph vertices $y,y'$ connected by an edge $(y,y')$: Delete edge
$(y,y')$, add a vertex $z$ and two edges, $(y,z)$ and $(z,y')$. The
transformation of elastic structure (change in the star list) is
induced by the change of topology, because the elastic graph is
primitive. This two--transformation grammar with energy minimization
builds {\it principal trees} (and principal curves, as a particular
case) for datasets. A couple of examples are presented on
Fig.~\ref{examples}. For applications, it is useful to associate
one-dimensional continuums with these principal trees. Such a
continuum consists of node images $\phi(y)$ and of pieces of
straight lines that connect images of linked nodes.

\section{Factorization and transformation of factors}

\begin{figure}
\centering{\includegraphics[width=80mm, height=30mm]{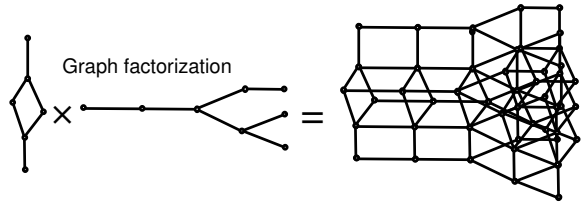} }
\caption{The Cartesian product of graphs. \label{FigFactor}}
\end{figure}

If we approximate multidimensional data by a $k$-dimensional object,
the number of points (or, more general, elements) in this object
grows with $k$ exponentially. This is an obstacle for grammar--based
algorithms even for modest $k$, because for analysis of the rule $A
\to B$ applications we should investigate all isomorphic copies of
$A$ in $G$. The natural way to avoid this obstacle is the principal
object factorization. Let us represent an elastic graph as a
Cartesian product of graphs (Fig.~\ref{FigFactor}). Cartesian
products $G_1 \times \ldots \times G_r$ of elastic graphs $G_1,
\ldots G_r$ is an elastic graph with the vertex set $V_1 \times
\ldots \times V_r$. Let $1 \leq i \leq r$ and $v_j \in V_j$ ($j \neq
i$). For this set of vertices, $\{v_j\}_{j\neq i}$, a copy of $G_i$
in $G_1 \times \ldots \times G_r$ is defined with vertices
$(v_1,\ldots v_{i-1}, v, v_{i+1}, \ldots v_r)$ ($v\in V_i$), edges
$((v_1,\ldots v_{i-1}, v, v_{i+1}, \ldots v_r), (v_1,\ldots v_{i-1},
v', v_{i+1}, \ldots v_r))$ ($(v,v') \in E_i$) and, similarly,
$k$-stars of the form $(v_1, \ldots v_{i-1}, S_k, v_{i+1},\ldots
v_r)$, where $S_k$ is a $k$-star in $G_i$. For any $G_i$ there are
$\prod_{j, j\neq i} |V_j|$ copies of $G_i$ in $G$. Sets of edges and
$k$-stars for Cartesian product are unions of that set through all
copies of all factors. A map $\phi: V_1 \times \ldots \times V_r \to
R^m$ maps all the copies of factors into $R^m$ too. {\it Energy of
the elastic graph product is the energy sum of all factor copies.}
It is, of course, a quadratic functional of $\phi$.

The only difference between the construction of general elastic
graphs and factorized graphs is in application of transformations.
For factorized graphs, we apply them to factors. This approach
significantly reduces the amount of trials in selection of optimal
application. The simple grammar with two rules, ``add a node to a
node or bisect an edge," is also powerful here, it produces products
of primitive elastic trees. For such a product, the elastic
structure is defined by the topology of the factors.

\section{Conclusion: adaptive dimension and principal cubic
complexes}

In the continuum representation, factors are one-dimension
continuums, hence, a product of $r$ factors is represented as an
$r$-dimensional {\it cubic complex} \cite{CubMatPol} that is glued
together from $r$-dimensional parallelepipeds (``cubes"). Thus, the
factorized principal elastic graphs generate a new and, as we can
estimate now, a useful construction: a principal cubic complex. One
of the obvious benefits from this construction is adaptive
dimension: the grammar approach with energy optimization develops
the necessary number of non-trivial factors, and not more. These
complexes can approximate multidimensional datasets with complex,
but still low-dimensional topology. The topology of the complex is
not prescribed, but adaptive. In that sense, they are even more
flexible than SOMs. The whole approach can be interpreted as a
intermediate between absolutely flexible {\it neural gas}
\cite{NeuralGaz} and significantly more restrictive {\it elastic
map} \cite{GorZinComp2005}. It includes as simple limit cases the
$k$-means clustering algorithm (low elasticity moduli) and classical
PCA (high $\mu$ for $S_2$ and  $\mu \to \infty$ for $S_k$, $k>2$).

\end{document}